\documentclass[letterpaper, 10 pt, conference]{ieeeconf}  

\IEEEoverridecommandlockouts                              

\usepackage{amsmath}
\usepackage{bm}
\usepackage{amssymb}
\usepackage{amsfonts}

\usepackage{color}
\usepackage{times}
\usepackage{caption}
\usepackage[pdftex]{graphicx}
\usepackage{amsmath}
\usepackage{amssymb}
\usepackage{subfigure}
\usepackage{latexsym}
\usepackage{hyperref}
\usepackage{siunitx}

\usepackage{dblfloatfix}

\usepackage[linesnumbered, ruled, vlined]{algorithm2e}  
\usepackage[capitalise]{cleveref}                       

\usepackage{todo}

\title{MaestROB: A Robotics Framework for Integrated Orchestration of Low-Level Control and High-Level Reasoning}

\author{Asim Munawar$^{1}$, Giovanni De Magistris$^{1}$, Tu-Hoa Pham$^{1}$, Daiki Kimura$^{1}$, Michiaki Tatsubori$^{1}$,\\ Takao Moriyama$^{1}$, Ryuki Tachibana$^{1}$ and Grady Booch$^{2}$ 
\thanks{$^{1}$IBM Research - Tokyo, Japan
        {\tt\small \{asim, giovadem, pham, daiki, mich, moriyama, ryuki\}@jp.ibm.com}}%
\thanks{$^{2}$IBM Research, USA
        {\tt\small gbooch@us.ibm.com}}%
}

\begin{document}

\maketitle
\thispagestyle{empty}
\pagestyle{empty}

\begin{abstract}
  This paper describes a framework called MaestROB.
  It is designed to make the robots perform complex tasks with high precision by simple high-level instructions given by natural language or demonstration.
  To realize this, it handles a hierarchical structure by using the knowledge stored in the forms of ontology and rules for bridging among different levels of instructions.
  Accordingly, the framework has multiple layers of processing components; perception and actuation control at the low level, symbolic planner and Watson APIs for cognitive capabilities and semantic understanding, and orchestration of these components by a new open source robot middleware called Project Intu at its core.
  We show how this framework can be used in a complex scenario where multiple actors (human, a communication robot, and an industrial robot) collaborate to perform a common industrial task.
  Human teaches an assembly task to Pepper (a humanoid robot from SoftBank Robotics) using natural language conversation and demonstration.
  Our framework helps Pepper perceive the human demonstration and generate a sequence of actions for UR5 (collaborative robot arm from Universal Robots), which ultimately performs the assembly (e.g. insertion) task.
\end{abstract}

\section{INTRODUCTION}
\label{sec:intro}
Making robot programming feasible for beginners and easy for experts is the key to use the robots beyond their conventional applications in industrial manufacturing and assembly.
Industrial and communication robots of the future will understand natural interfaces like speech and human demonstrations to learn complex skills. 
Instead of a single brain-like component, these robots will be powered by many smaller cognitive services that will work in harmony to exhibit a higher level of cognition.
Each service in the system will demonstrate basic intelligence to efficiently perform a single well-defined task.
A framework for connection and orchestration of these services will be the key to a truly intelligent system.
Although often used interchangeably, we use the term ``cognitive'' instead of ``machine learning'' or ``artificial intelligence'' to give a notion of human like intelligence across multiple domains.
Mostly, the term artificial intelligence is used very broadly, while machine learning usually refers to systems that solve a distinct problem in a single domain.

To achieve this higher level of cognitive capabilities, in this paper we present a robotics framework -- MaestROB. 
Different components of MaestROB communicate through a novel robotics middleware named Project Intu.
Intu is provided as an open source project at \mbox{\url{https://github.com/watson-intu/self}}.
Intu uses IBM Watson APIs~\cite{url:watson} to provide a seamless access to many services including conversation, image recognition etc.
With Intu at its core, MaestROB introduces a hierarchical structure to planning by defining several levels of instructions.
By using the knowledge and ontology of physical constraints and relationships, these abstractions allow the grounding of human instructions to actionable commands. 
The framework performs symbolic reasoning at higher level, which is important for long term autonomy and to make the whole system accountable for its actions.
Individual skills are allowed to use machine learning or rule based systems.
We provide a mechanism to extend the framework by developing new services or connecting it with other robot middlewares and scripting languages.
This allows the higher level reasoning to be done using PDDL (Planning Domain Definition Language) planner with the proposed extension for handling semantics, while the lower level skills can be executed as ROS (Robotics Operating System) nodes.
In MaestROB the primitive intelligence of each component is orchestrated to demonstrate complex behaviors.
Important features of MaestROB include but are not limited to a cloud based skill acquisition and sharing service, task planning with physical reasoning, perception service, ability to learn from demonstration, multi-robot collaboration, and teaching by natural language.

We show the capabilities of the framework by a scenario where a human teaches a task to a communication robot (Pepper~\cite{url:pepper}) by demonstration.
The robot understands the tasks and collaborates with an industrial manipulator (UR5~\cite{url:ur5}) to execute the task, using the action primitives that UR5 has previously acquired by learning or programming.
The industrial robot has the ability to perform physical manipulation, but it lacks the key sensors that can help in a particular situation (e.g. error recovery etc.).
Using the planning and collaboration services provided by MaestROB, the communication robot having these sensors can analyze and convert the plan to a command sequence for the robotic arm.
The learning capabilities of the framework and the hierarchical control capabilities of the middleware help to enable these tasks easily without the need of any explicit programming.

\section{RELATED WORK}
\label{sec:relatedwork}

Middleware serves the purpose of gluing together various components of the robot and communication between them.
Robotics Operating System (ROS)~\cite{conf:icra:quigley2009} is arguably the most used robotics middleware especially in the research community.
Although ROS provides a backbone structure for many nodes to collaborate, it does not intrinsically provide any components that can help in planning or training the robots to perform a task. 

Some other papers discuss smaller frameworks to solve more complex problems.
ROSPlan~\cite{conf:icaps:cashmore2015} is a task planning system that can generate a sequence of actions to achieve the goal when the initial model of the environment is known.
ROSPlan being a planning system depend on other components for learning and communication.
ROSPlan is not a full robot framework, also it does not provide any mechanism to train the robots.

The problem of providing the robot with reasoning capabilities in order to allow them to generate new motion types autonomously has been tackled throughout the recent years.
The RoboHow project aimed at enabling robots to perform everyday manipulation activities from web navigation, and human observation.
The project introduced KnowRob, a knowledge base on actions, objects, properties and relations~\cite{article:ijrr:tenorth2013}.
This knowledge base was combined with object recognition algorithms to recognize and reason on visual observations through the RoboSherlock software framework~\cite{conf:icra:beetz2015}.
In the RoboBrain knowledge engine, the nodes of the graph structure can represent any type of robotic concept (e.g., grasping features, trajectory parameters and visual data)~\cite{article:corr:saxena2014}.
The RoboBrain knowledge engine was used to execute manipulation tasks from instructions given in natural language~\cite{article:ijrr:misra2016}.
As RoboBrain uses the knowledge acquired from semi-reliable sources, the outcome actions of an instruction are not predictable.
We on the other hand wants to communicate using natural language but at the same time we would like the end actions to match our expectations.

An increasing number of research papers use machine learning and cognitive technologies for collaboration, handling uncertainties, and taking optimal actions.
We discuss some of the interesting research in the area of cognitive robots with capabilities beyond that of the current generation of robots.

S. Levine et al.~\cite{article:ijrr:levine2017} defined a hand-eye coordination for robotic grasping using monocular images.
It shows how deep convolutional neural networks can be used with big data to learn a complex task in an uncertain environment without explicit need of camera calibration and the robot pose estimation.
Delft team won the Amazon picking challenge 2016~\cite{arxiv:carlos2016} by using machine learning for pose estimation, grasp planning, and motion planning.
T. Inoue et al.~\cite{conf:iros:inoue2017} showed how the data from conventional sensors can be combined with the deep reinforcement learning to solve precision insertion task.
A. Munawar et al.~\cite{conf:wacv:asim2017} presented an anomaly detection system using vision.
Such systems can be used to enable the systems to keep a check on themselves with little intervention from the humans.
J. Connell et al.~\cite{conf:agi:connel2012} presented a system for physical manipulation with a robot arm.
The system can be used to teach the robot to learn new actions using a a fixed grammar.
The grammar is basic and can be difficult to extend to make it more general.
R. Paul~\cite{conf:ijcai2017:paul} presents a probabilistic model that enables incremental grounding of natural language utterances using learned knowledge from accrued visual observations and language utterances.
The model infers over the constraints for future actions in the context of a rich space of perceptual concepts.

While interesting, most of the existing research use customized learning and control solutions that cannot be generalized.
The proposed framework is an effort to present a common framework that can handle all these scenarios and beyond without any redundant effort.

\begin{figure}[t]
\centering
\includegraphics[width=1.0\columnwidth]{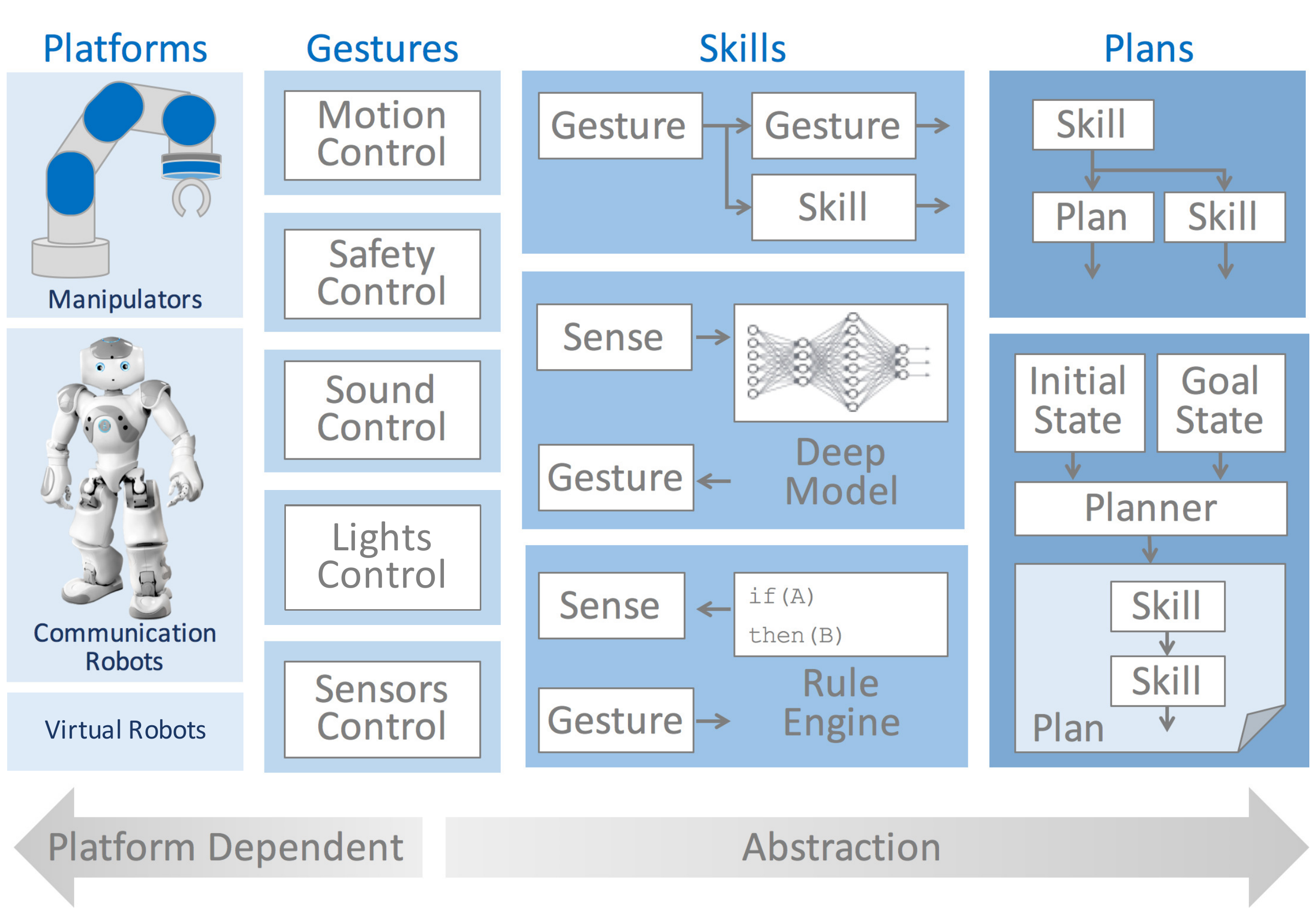}
\caption{Different level of instructions in MaestROB.}
\label{fig:cmd-act-skill}
\end{figure}

\begin{table}[b]
\centering
\caption{Example of instructions at different levels.}
\label{table:cmdtable}
\includegraphics[width=1.0\columnwidth]{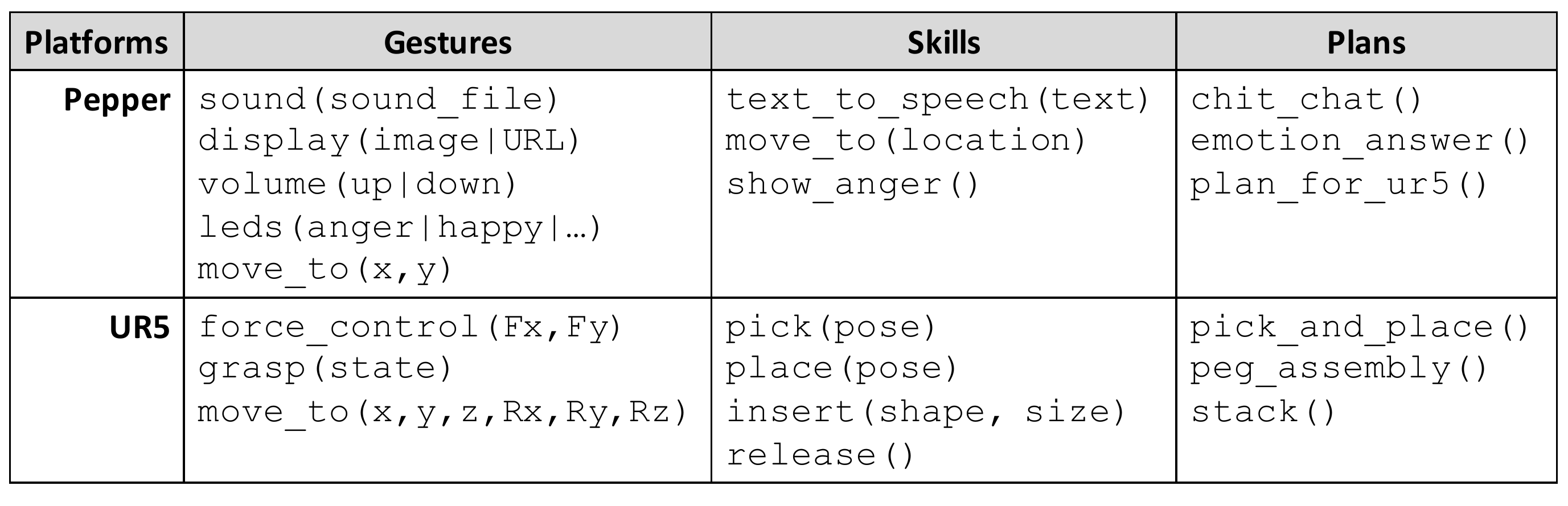}
\end{table}

\section{MOTIVATION AND CONCEPTS}
\label{sec:maestrobmot}

The motivation behind MaestROB is to create a framework that performs accurate physical manipulations in the real world by watching demonstrations or taking natural language instructions.
Humans communicate at higher level of abstraction and assume the underlying knowledge. 
Machines on the other hand cannot accurately comprehend such vague instructions.
In order to make the robots understand natural language commands in a predictable manner, we propose a hierarchical structure of instructions.
Fig.~\ref{fig:cmd-act-skill} shows different levels of instructions in MaestROB, namely gestures, skills, and plans.
Gestures are platform dependent but abstraction is provided as we move to higher level instructions.
Examples of different level of instructions are given in Table~\ref{table:cmdtable}.

\subsection{Gestures}
Gestures are used to directly control the robot.
They are equivalent to the motor skills in a human body.
Gestures are executed by the underlying platform or the robot controller.
Depending on the platform, some of the gestures might not be available.

\subsection{Skills}
The next level of abstraction is the skills.
We define a skill as a piece of logic that can consume sensor input and generate a list of gestures.
A skill is an atomic operation that performs a part of the overall task.
Skills however cannot be executed on their own.
MaestROB provides three methods of teaching new skills to the robot:

\begin{enumerate}
\item \textit{List of gestures or skills}:
  In its simplest form, a skill consists of sequential or parallel list of gestures or other skills.
  For example, a pick skill can take the pick position and perform the sequence of gestures by going on top of the position, then going down, and finally closing the gripper.
\item \textit{Rule based}:
  In this method, a set of rules can be defined to consume the sensor's input and to issue appropriate gestures.
  Rule based skills are simply defined as ``if\{A\}then\{B\}'' rules.
  One example can be to stop the robot if the end-effector position is outside a predefined safety zone.
\item \textit{Machine learning}:
  Some skills are too difficult to be defined by a program.
  MaestROB provides a cloud based learning API that can be used to learn complex skills like different shape insertion task or visual servoing.
  The learning service supports supervised and reinforcement learning paradigms.
  The inference is done locally to satisfy real-time constraints of the robots, while the models are stored and learned in the cloud.
  In this manner, robots with minimal processing capabilities can also learn new skills by making REST calls to the MaestROB learning API.
  The API also supports transfer learning of models from a simulator to the real environment.
\end{enumerate}

MaestROB provides an interface to extend the framework by implementing additional methods to define or learn new skills.
A skill can take one or more arguments as its input, e.g. \texttt{place(pose)}, \texttt{moveTo(pose, pose)}.
All skill return \texttt{success} or \texttt{failure}, based on the results of actual execution of the action in the prescribed amount of time.

\subsection{Plans} 

A plan is conceptually a high-level abstraction for going from a given initial state to a goal state.
It consists of a sequence of skills that are defined, learned, or computed by a symbolic planner.
A plan corresponds to a single instruction in a user manual or a single command issued by human.
The role of the planner in the framework is to act as a bridge between the cognitive semantics and the skill.
The framework currently provides three methods for defining new plans.

\begin{enumerate}
\item \textit{List of skills or plans}:
  A plan can be defined as a parallel or sequential list of skills or other plans defined either programmatically or by using natural language communication using fixed grammar.
  This method is inspired by previous work by J. Connell et al.\cite{conf:agi:connel2012}, etc.
  For example, we can tell the robot that he can learn to wave hands by raising the right hand and moving it right and left a few times.
\item \textit{Learning from demonstration or conversation}:
  Plans can also be described by defining the final or goal state verbally or programmatically.
  MaestROB provides a mechanism for grounding the natural language commands to the intended goal states by using natural language classification and mapping.
  Another method supported by the framework is to show the key states of the system, instead of giving the initial and the goal states explicitly.
  Humans can understand the sequence of actions they need to perform just by looking at the initial and the final state of a task.
  By using the perception and relationship extraction, the planner can help the robot do the same by using the available skills.
  Before computing and executing the plan, the framework checks the current condition of the world to know the initial state.
  
\end{enumerate}
Similar to skills, we define an interface to implement other methods of planning.
A plan is a logic that can generate a sequence of one or more skills.
All plans return \texttt{success} or \texttt{failure}.

Unlike skills, a plan can be executed independently.
It can consist of a single skill, a list of skills or dynamically computed skills based on a planner or a similar system.
Plans can be launched by human instructions, in response to some condition, or in a continuous loop.
In case of a skill failure a new plan can be generated to accomplish the target by using current state as the initial state.
However, human intervention may be required, if the planner fails to find a suitable plan.

\begin{figure}[t!]
  \centering
  \includegraphics[width=1.0\columnwidth]{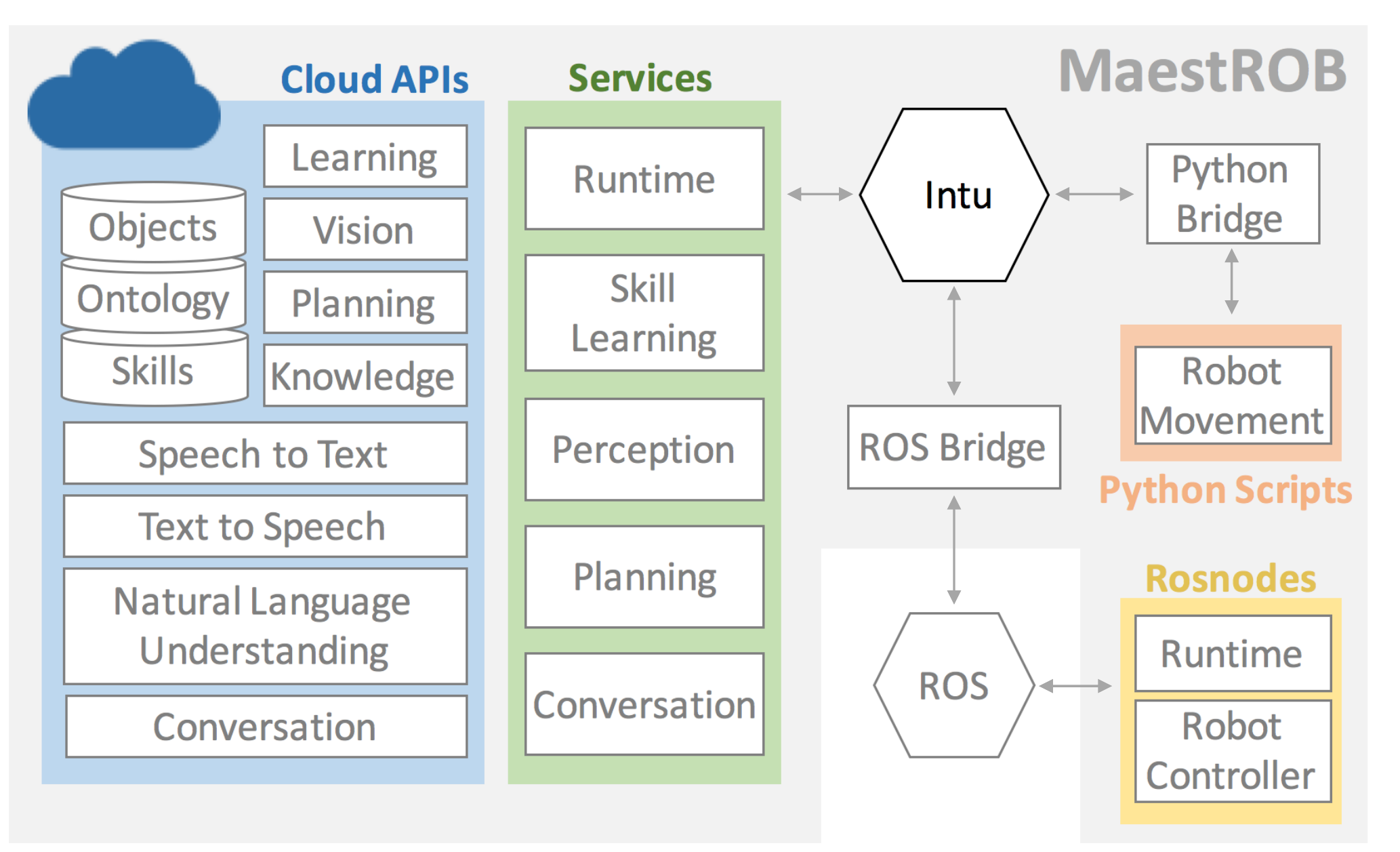}
  \caption{MaestROB framework.}
  \label{fig:first}
\end{figure}

\section{MaestROB FRAMEWORK}
\label{sec:maestrobframe}

MaestROB is a scalable and extendable framework that can be used to control robots or smart environments.
It is built using a hybrid architecture, encompassing explicit symbolic computation at the center and neural networks at the edge.
At the core of the framework is the robot middleware Intu.
The library around the middleware consists of numerous algorithms and components, each of which performs a well defined task.
In this section, we discuss the middleware and the key services of MaestROB.

We provide numerous services to enable a large number of commonly occurring use case scenarios.
A service is defined as a set of components that performs a well defined complex task, for example conversation.
Components can be shared among multiple services.
We use the term components loosely to represent data storage, program, learned models or any part of the framework.
Fig.~\ref{fig:first} shows an abstract level configuration of the middleware, cloud APIs, and the main services of the framework.

\subsection{Project Intu}
Project Intu is provided as an open source platform for embodied cognition\footnote{\url{https://github.com/watson-intu/self}}.
It is based on a cognitive architecture called Self.
Self is an agent-based architecture that combines connectionist and symbolic models of computation, using blackboards for opportunistic collaboration.
Project Intu provides a mechanism for connecting and orchestrating cognitive services in a manner that brings higher level cognition to an embodied system.

Self is inspired by Minsky's Society of Mind~\cite{book:minsky:1986}, therefore, behavior takes place in the context of multiple concurrent agents who communicate opportunistically via blackboards.
Inspired by Brook’s subsumption architecture~\cite{article:ijra:brook1986}, behavior takes place in a hierarchy of cognition, from involuntary reflexes to voluntary skills to goals and planning.

Self imposes a clear separation of concerns among perception, actuation, models, and behavior, and as much as possible, behavior is either taught or is learned, not programmed.
For perception the extractors are used to process the raw sensor streams into a meaningful data (e.g. speech to text).
The data then goes to the respective classifiers or agents for further processing.
Actions are performed by learning or programming different skills.
Based on a goal that needs to be achieved, the plan manager finds the most suitable execution plan before invoking that plan (sequence of skills).
Most of the components communicate opportunistically by using the publish and subscribe (pub/sub) model of Intu.
The communication with the outside world or other instances of Self happens through topics.
Plugins can be developed to connect Intu with other middleware’s and scripting languages.

Open source version of Intu comes with many components that allow a seamless access to IBM Watson services~\cite{url:watson}.
Intu is devised to be applicable to a multitude of use cases, from avatar to concierge to retail to elder care to industrial robots.
It is available for a number of platforms, including Linux, Windows, macOS, Nao, Pepper, and Raspberry Pi.

\setcounter{figure}{3}
\begin{figure*}[b]
\centering
\includegraphics[width=1.0\textwidth]{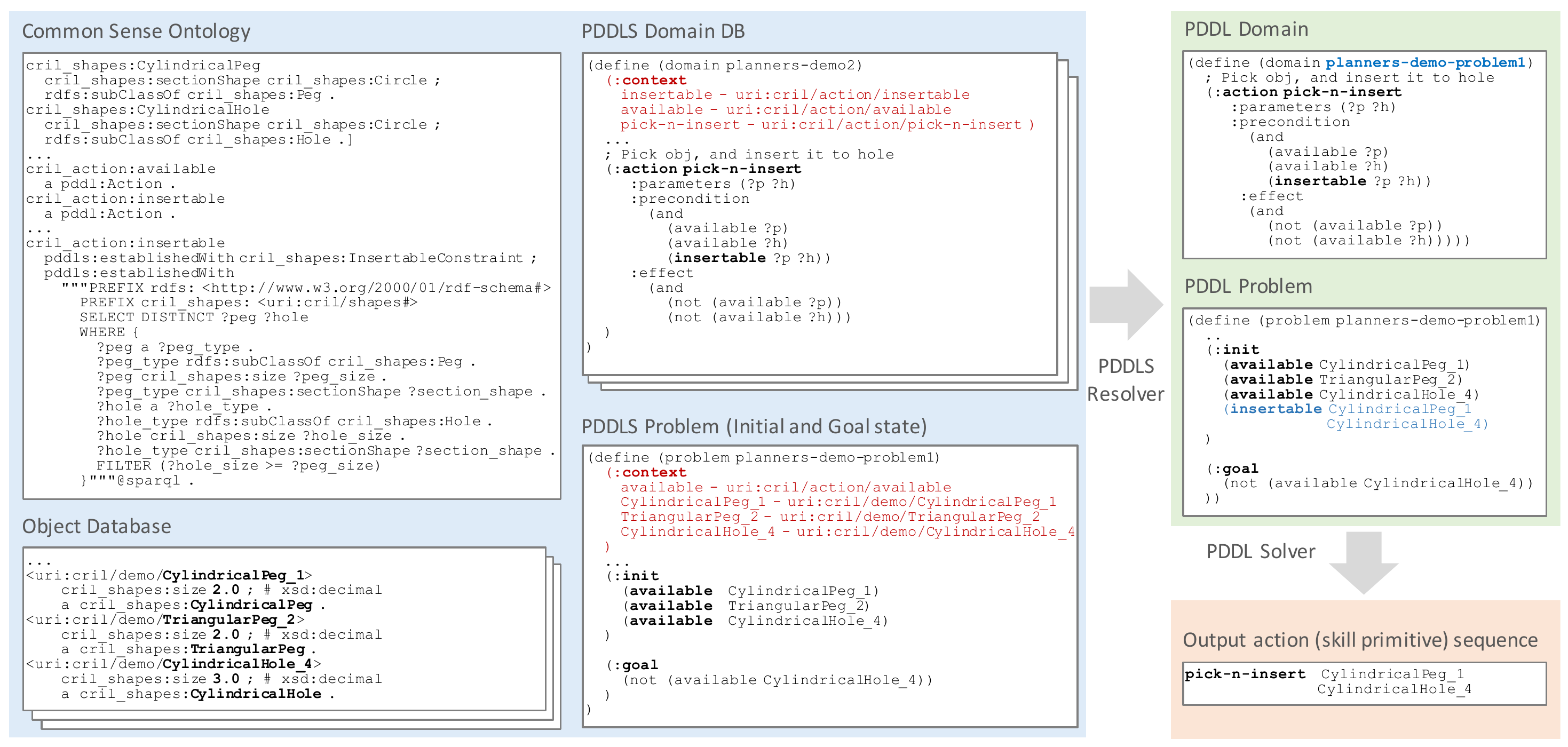}
\caption{An example to show how proposed PDDL with semantics can leverage the ontology and reasoning to solve a problem.}
\label{fig:pddls}
\end{figure*}

\subsection{Perception Service}
Sensing and perception is required for both physical manipulation and learning. 
MaestROB supports different sensors including microphone, camera, and touch.
Raw stream from the sensors is converted into meaningful format by the help of extractors and classifiers.

While different sensors are used by MaestROB, in this section we will focus on the vision sensor (camera).
Camera is used for recognizing a human in front of the robot, estimating the age and gender, recognizing objects and their poses etc.
For accurate visual perception in industrial settings, we assume that an object database provides the properties (shape, size etc.) of all the objects of interest.
Visual perception computes the pose for each instance of the object in the visible world.
For the demonstration, we have used a barcode based pose estimation technique proposed by H. Kato~\cite{kato2002artoolkit}.

In addition to perception, spatial relationships are also extracted geometrically to understand the state of the physical world.
Relationships currently supported by the framework are in, on, right, left, front, and back.
With the supported interface of object detection and relationship extraction, the framework can be extended by including better pose estimation algorithms.

\subsection{Converstaion Service}
MaestROB conversation service uses IBM Watson cloud based Conversation API~\cite{url:watson}.
The Conversation API combines machine learning, natural language understanding, and integrated dialog tools to graphically create conversation flows. 
MaestROB conversation service helps the machine understand human intentions and act accordingly.
Depending on the conversation model, the robot can clarify missing information or even start a dialog proactively.
This is especially important for robots working as salesman or concierge.

Raw stream from microphone is extracted by \textit{text extractor} that calls IBM Watson Speech-To-Text (STT) API.
The text then goes to a \textit{natural language classifier} which internally calls IBM Conversation API and classifies the text according to its intention.
For example, in case of a \textit{question intent}, the \textit{question agent} finds the appropriate response and generates an \textit{answer goal}.
The \textit{goal agent} then executes the appropriate plan with the help of \textit{plan manager}.
The \textit{plan manager} may decide to invoke \textit{say} action, which is ultimately invoked by \textit{skills manager} by calling the appropriate \textit{skill}.

\setcounter{figure}{2}
\begin{figure}[t]
\centering
\includegraphics[width=1.0\columnwidth]{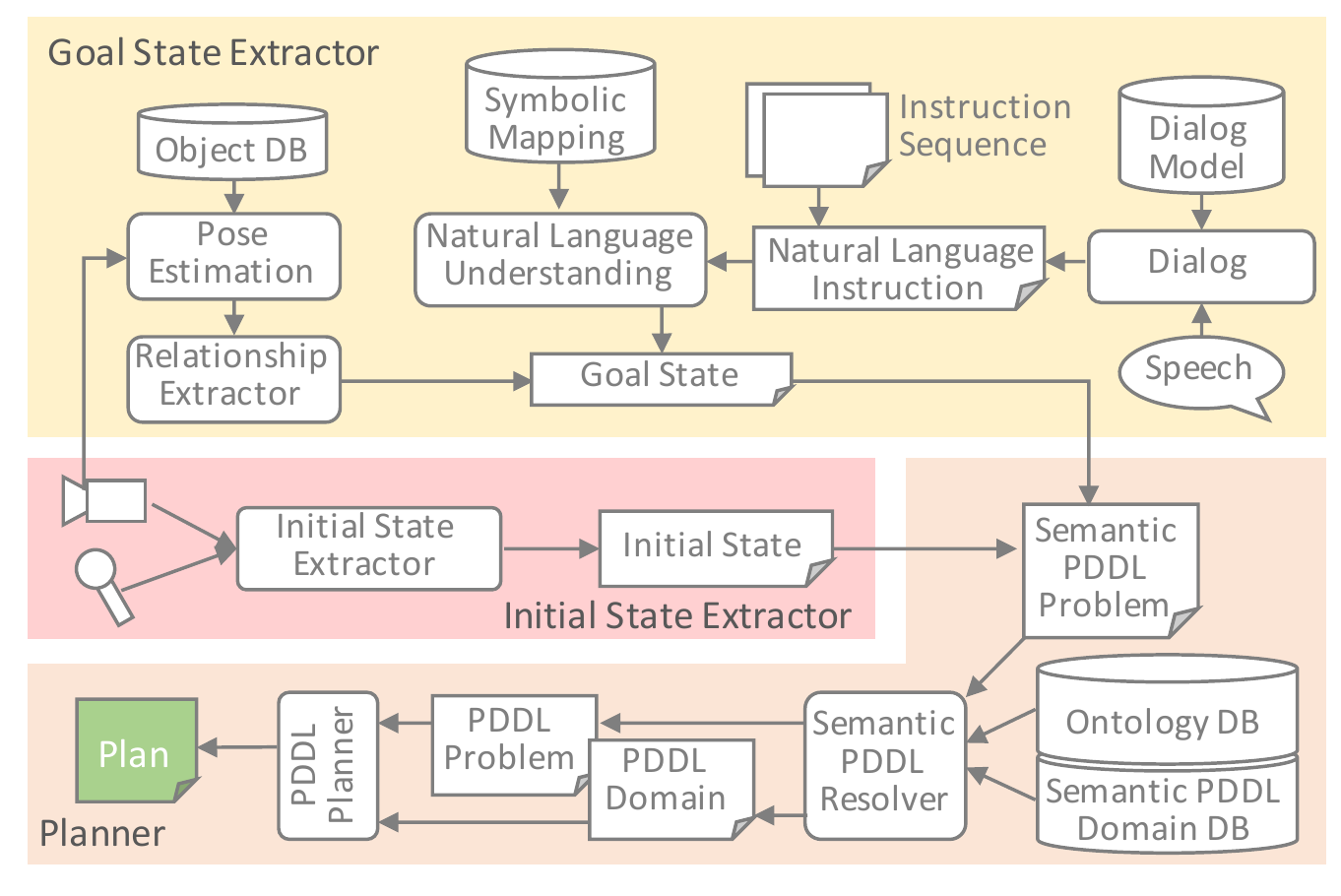}
\caption{MaestROB Planner.}
\label{fig:planner}
\end{figure}
\setcounter{figure}{4}

\subsection{Planning Service} 
Planner is responsible to observe the current state of the system, understand the goal and generate a sequence of skills to achieve the goal.
Configuration of the proposed planner is shown in Fig.~\ref{fig:planner}.

Initial state of the world is extracted by the ``initial state extractor''.
The initial state is the truth about the world and is always detected before computing a plan.
This is usually done by using vision sensors.
Other input devices like mic or depth sensors can also be used to define the initial state.
The first step in defining the initial state is to get the poses (positions and orientations) of all the available objects.
We assume that all the objects (classes) are defined in an object database.
Computing poses of all the class instances is not enough, we also need to define the states of the objects, e.g. if a hole is filled or not.
This is done by relationship extractor of the perception service. 

Understanding the goal state is also a crucial part of the planner.
The goal can be defined either by a key frame based demonstration or by using natural language. 
In MaestROB we use symbolic mapping for grounding natural commands to the respective goal state.
Instead of simple matching, Watson Natural Language Classification API~\cite{url:watson} is used to match a command to the intended goal state.
Although this may limit the variation of natural language that can be used, it produces consistent and predictable results.

For the planning part, we extended PDDL~\cite{tech:ghallab1998} (Planning Domain Definition Language) for describing and indexing skills by allowing semantic annotations.
We call the extended version of the language as PDDLS (Planning Domain Definition Language with Semantics).
The primary role of the semantics is to give the compatibility among skills and the recognized states.
Symbols are bound to be globally identifiable references as URIs (Uniform Resource Identifier), for which equals-to, is-a, and other equality relationships and compatibilities are defined in standard ontologies.
Another important role of the semantic resolution provided by the framework is to use ontology to compensate for the constraints, or common sense in the domain, which is difficult to be captured through cognition.
For example, while an image analysis can recognize the shape and size of objects, it would not directly give the constraints for combining those objects - e.g. if a peg can be inserted into a hole or not.

The skills in the skill database can be defined as actions in PDDLS for a particular domain, in which the preconditions and effects of the skills are described.
A planning query is given as a PDDLS problem, in which the goal and initial states are described as PDDL with semantic labels.
Extended semantic annotations enable global linking between actions and problems with semantic resolution.
The mechanism for the semantic resolution is beyond the scope of this paper.
The semantics, actions, and required relationships are defined in PDDLS domain files, therefore, a suitable domain file must be defined explicitly by the user to generate an optimal plan for the problem.
The output of the planner is a sequence of skills that is executed by the runtime, which might request the planner to generate an alternative plan in case of a failure.

Fig.~\ref{fig:pddls} shows a simple example to illustrate how the planner works.
We have two pegs and a hole and only the cylindrical peg can be inserted in the hole.
The common sense ontology is provided to the system, including the knowledge for obtaining constraints among objects, such as whether an object can be inserted (insertable) to another object or not.
We search for the appropriate domain descriptions in the PDDLS domain database.
The domain file defines, all the available actions, which in this case include only one action ({\it pick-n-insert}).
Note that in the context section symbols are semantically annotated by URIs (shown in red color), which are defined in the ontology or runtime object properties.
The initial and goal states of the PDDLS problem file, as well as runtime object properties ontology (e.g. object shapes and sizes), are defined by perception and grounding of natural language commands as defined above.
The PDDLS resolver then generates a runnable PDDL problem and domain file, using the semantic annotations to resolve necessary constraints.
These constraints are shown in blue color in the figure and they are required by the PDDL solver (PDDL planner) to find a valid solution.
The problem and the domain files are used by the PDDL solver to output the correct sequence of actions, which in this case is to perform the ``pick-n-insert'' to put the cylindrical peg in the cylindrical hole.

\subsection{Skill Acquisition}
The process for acquiring skills is different from that of plans.
Plans can be acquired using natural language communication or demonstration, provided that all the skills are available in the skill database.
The skills are to be acquired individually by programming, defining rules, or machine learning.
MaestROB provides cloud based machine learning API to allow the robot to acquire or fine tune machine learning based skills.

\subsection{Bridges}
MaestROB can be connected to other scripting languages, robot middlewares, or rule engines by creating bridges.
Bridges are currently available for ROS, and Python.
A bridge converts the blackboard based communication that is native to Intu into a protocol that is understandable by other middlewares or languages.
ROS bridge, therefore, is an Intu agent and a ROS node at the same time.

\subsection{Runtime}
Runtime is the part of MaestROB that is responsible to execute a plan.
The runtime essentially takes the output of a planner which is a list of skills and execute them in the given sequential or parallel order.
In case of a failure in a skill, the runtime returns the feedback to the planner.
Then depending on the planner, a new plan can be generated as an alternate sequence of actions.
While the goal does not change, the current state of the world has to be recomputed.
If the system finds a situation where no suitable plan can be found, it may request help from the user.

\begin{figure}[b]
\centering
\includegraphics[width=0.8\columnwidth]{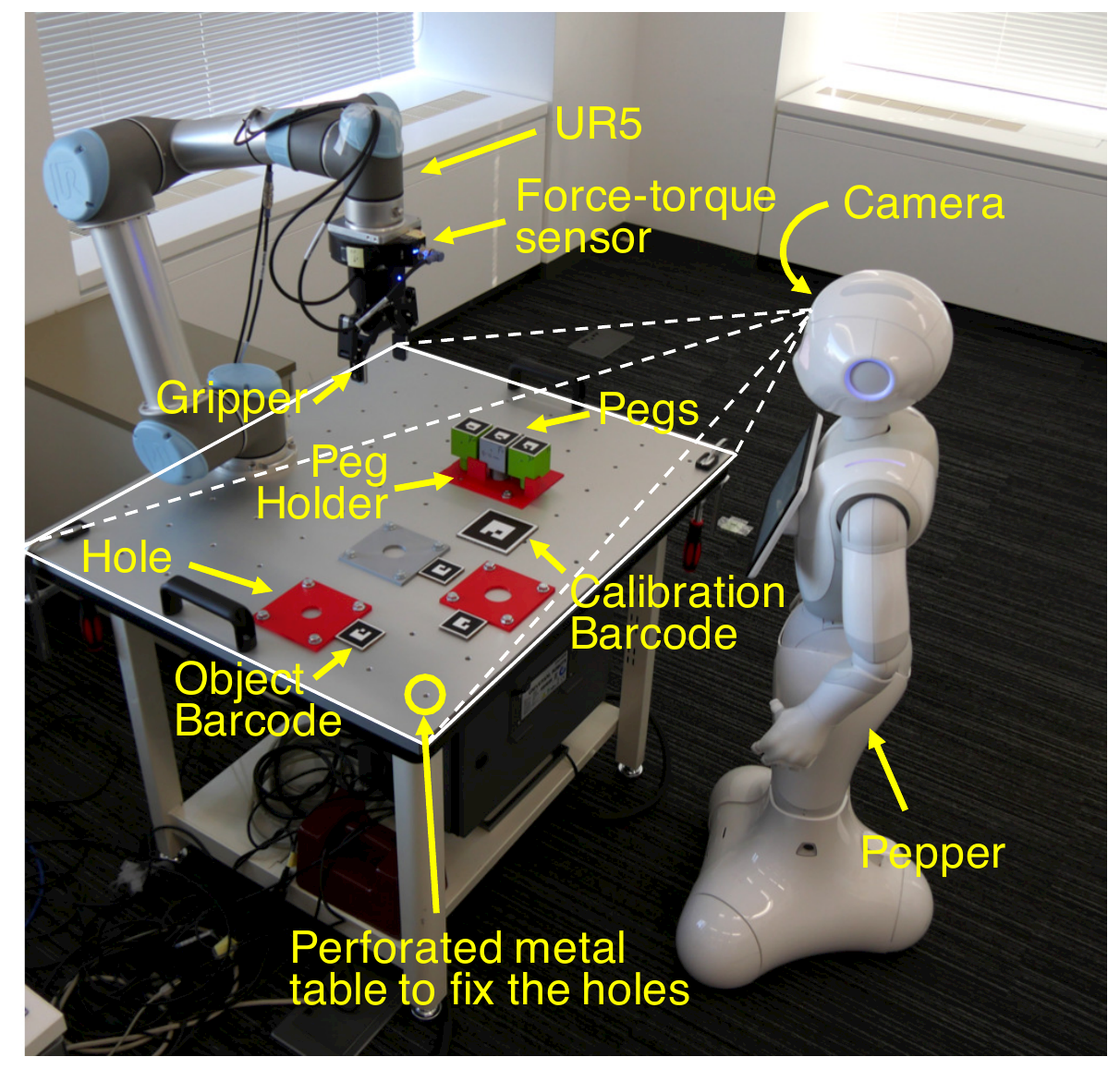}
\caption{Demo setup: Pepper uses its camera to analyze the location of the objects, creates a plan and send it to UR5.}
\label{fig:setuppic}
\end{figure}

\section{DEMONSTRATION}
\label{sec:demo}

In the demo scenario, two robots (UR5 and Pepper) collaborate with a human to perform a task.
MaestROB is general enough to handle both the communication robot and the industrial manipulator arm, we show how very different robots can achieve a common goal by using their strengths.

Pepper is a humanoid robot from Softbank Robotics that possesses several sensors including vision.
It is mobile, but it lacks a powerful gripper and cannot perform accurate physical manipulations.
On the other hand, UR5 (collaborative robot arm from Universal Robots) is a fixed industrial grade manipulator robot.
UR5 can do physical manipulations with high precision and repeatability ($\pm$\SI{0.1}{\milli\metre}) but it moves blindly due to the lack of any vision sensor.

In the demo scenario, the human is responsible for teaching the task by performing demonstration, controlling the robot, and taking final decisions.
The setup of the two robots for the demonstration is shown in Fig.~\ref{fig:setuppic}.
The demo can also run on other robot platforms, given the required gestures are available.

\begin{table}[]
\centering
\caption{Step-by-step demo scenario. Actors are human, Pepper, and UR5.}
\label{table:demosteps}
\includegraphics[width=1.0\columnwidth]{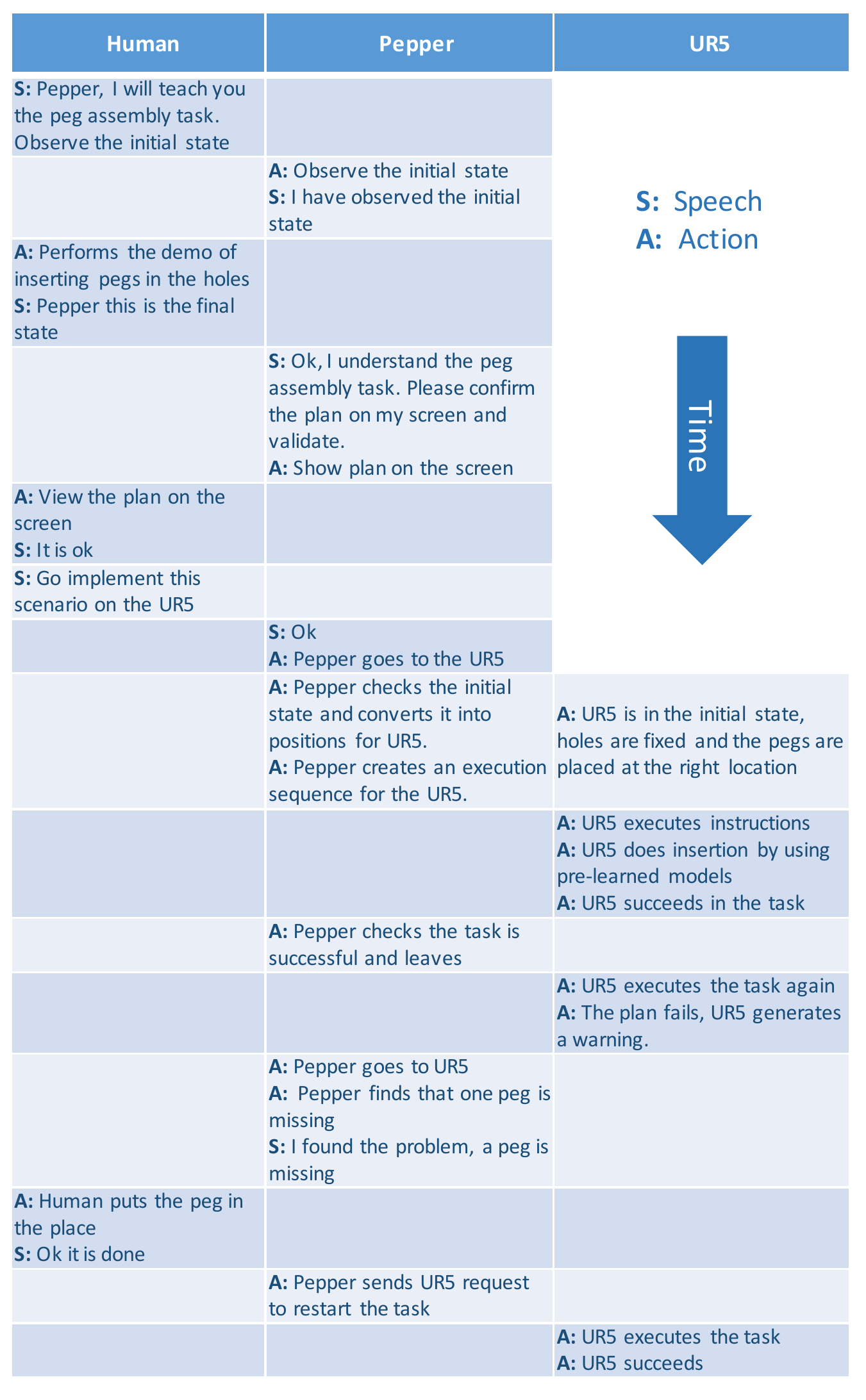}
\end{table}

Table~\ref{table:demosteps} shows the detailed scenario and the actions performed by different actors.
One of the strengths of MaestROB is the connection with IBM Watson APIs.
This allows to easily implement complex conversation systems.
Grounding of instructions given in a smooth natural language conversion to complex but predictable sequences of actions demonstrate the strength of MaestROB.

\begin{figure}[b!]
\centering
\includegraphics[width=0.96\columnwidth]{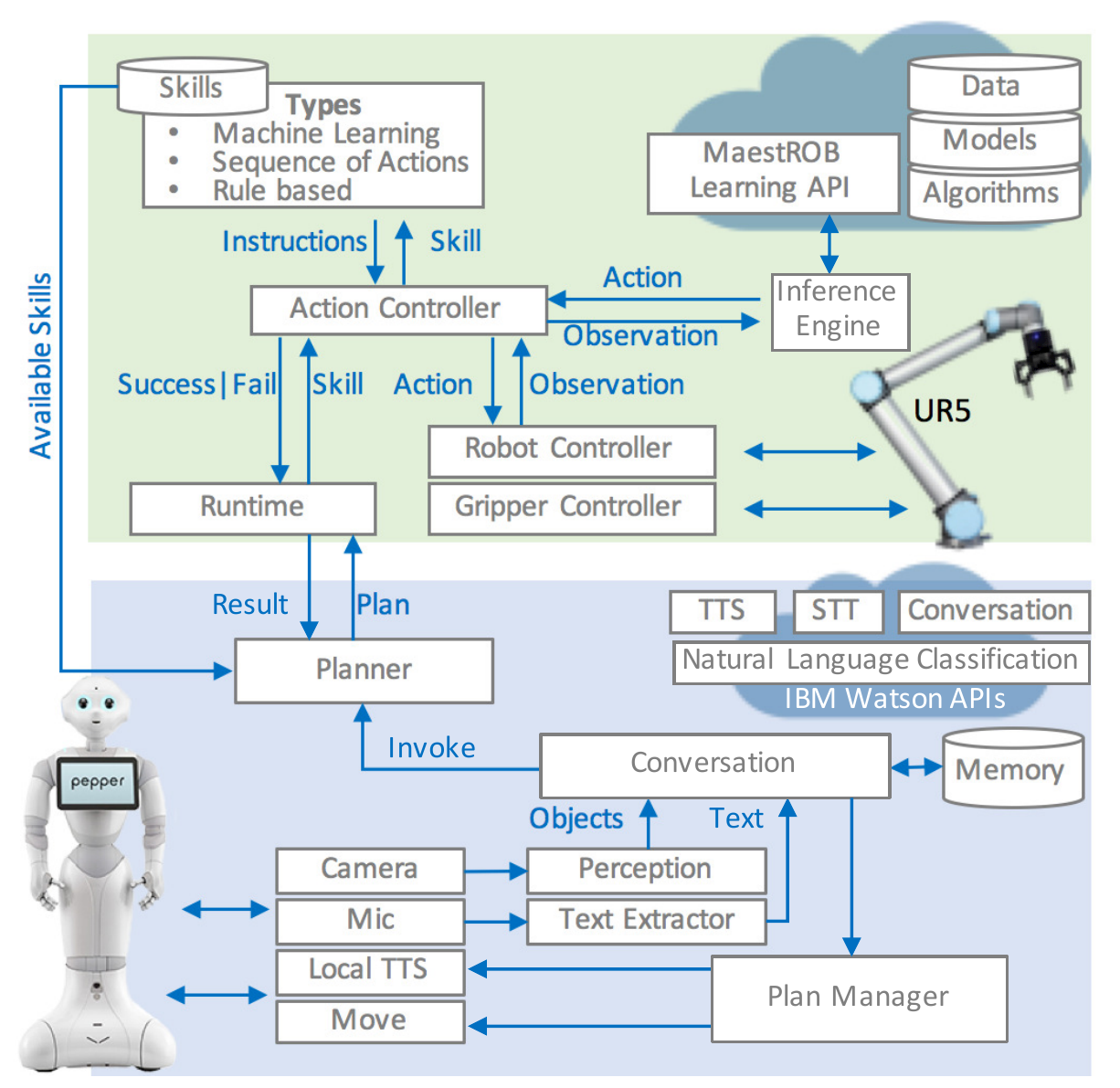}
\caption{Important services to run the demonstration. The services communicate via pub/sub model of Intu. Robot controller of UR5 is a ROS node and is connected with Intu via ROS bridge.}
\label{fig:arch}
\end{figure}

The demo starts with a human performing the task and having conversation with Pepper robot at the same time.
Understanding the intent of a command by using conversation service, the robot can understand when to capture key frames and when the demonstration is over.
In the example, the robot records the initial state and the final state of the demonstration.
From the conversation, it remembers the name of the task it is learning to perform (peg assembly task).
It also understands that the final state is the last frame of demonstration.
The initial and final frame are sent to perception service that uses barcode pose detection to detect the location of all the barcodes.
The barcode number of each part and the transformation between the barcodes and the objects are defined separately in the object database.
The state of the final frame is determined by the relationship extractor.
As the domain for the task is predefined to be insertion, the appropriate ontologies and the relationship database are loaded.

\begin{figure*}[t!]
\centering
\includegraphics[width=1.0\textwidth]{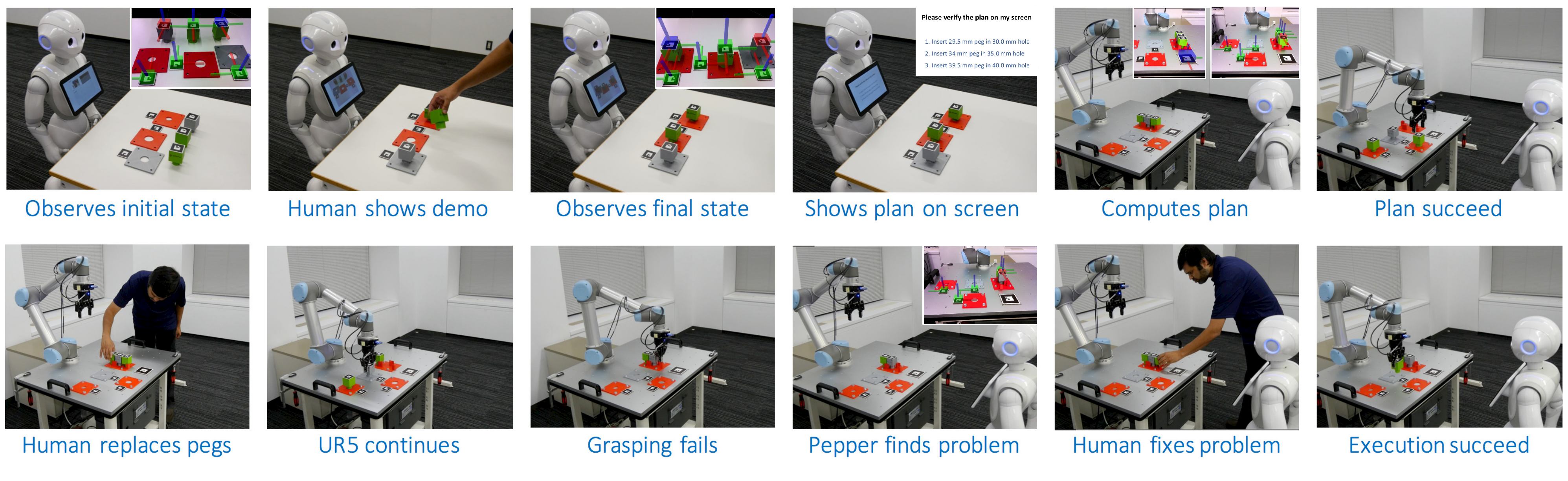}
\caption{Screenshots of the demo video (video is available at \url{https://www.youtube.com/watch?v=19JsdZi0TWU}). Barcode pose estimation results are shown as overlays.}
\label{fig:movie}
\end{figure*}

In this scenario, the goal is not used by Pepper to compute a plan for itself, rather Pepper implements the scenario on UR5.
Pepper uses predefined locations of the demo table and UR5 to move from one place to the other.
When Pepper arrives at UR5, it captures an image of the initial state.
This image is used to calibrate Pepper's camera w.r.t UR5 robot, the barcode pose location for all the objects are also computed.
Pepper uses the initial state observed from the image and the goal state learned from the human demonstration to generate an executable plan for the manipulator robot.
However, before this can be done, the skill database of UR5 is shared with Pepper.
The common sense ontology is used by the planner to check if an operation is permitted or not.
For example, it is not permitted to insert a big peg into a smaller hole.
After the plan is transmitted to UR5, it starts doing the task based on position control.
Once the task succeeds Pepper can return while UR5 continues to perform the task.
In this demo we have human helper to put the pegs back to the initial state before every UR5 iteration.
In a factory environment this is usually done by conveyer belts or other machines.

UR5 keeps executing the task unless something goes wrong, in this case UR5 warns about the failed plan.
In this case, Pepper goes back to the manipulator to observe the state of the world.
Pepper compares the current state with the initial state it had seen before. 
It finds that a peg is missing and raises a request for human assistance.
The human then comes to put the missing peg.
UR5 can resume to do the repetitive task of putting pegs in the respective holes.

Contrary to several simple skills that UR5 can perform, the insertion skill uses machine learning to find the direction of hole by using force-torque sensor.
It was trained by reinforcement learning using the MaestROB cloud based learning API.
The training method for the insertion skill is similar to T. Inoue et al.~\cite{conf:iros:inoue2017}.

Fig.~\ref{fig:arch} shows the MaestROB services that are running on each robot to accomplish the tasks defined in the demo scenario.
Most of the services use cloud based APIs to solve the problem. 
It is important to note that in order to make a plan for UR5, Pepper robot must be aware of the skills available for UR5 robot.
This is done by sharing UR5 skill database with Pepper.
In the demo scenario, Pepper is running an instance of the Project Intu middleware.
Intu makes the implementation of Speech-To-Text (STT), Text-To-Speech (TTS), conversation, perception, planning etc. easy and streamlined.
UR5 is running the plan generated by Pepper using MaestROB services on ROS.

Snapshots of some of the key moments in the demo video can be seen in Fig.~\ref{fig:movie}.
The demo video is available at \url{https://www.youtube.com/watch?v=19JsdZi0TWU}.

\section{CONCLUSIONS}
\label{sec:conclusions}

We present a framework to support the next generation of robots to help solve the problems that are not solvable by conventional programming methods.
The robot middleware (Project Intu) presented in this paper is now available as an open source project.
We also presented several key services that enable us to demonstrate sophisticated scenarios involving collaboration between multiple robots and human.
MaestROB is especially useful for small and medium-sized enterprises (SMEs), which need relatively quick time to market, frequent changes in manufacturing lines and have low production volumes.
The workers can communicate with the robot in natural language and teach it new skills or execute existing skills.

As a future direction, we would like to make the machine learning based skills sharable among multiple robots.
As mentioned in the paper and the demo, a failed plan usually requires human assistance.
One of the future directions can be to make the robot resolve common problems on its own.
We also plan to demonstrate a system that understands written and spoken instructions to create a complex object like IKEA furniture.



{\small
\bibliographystyle{ieee}
\bibliography{latex}
}

\end{document}